\documentclass[letterpaper, 10 pt, conference]{ieeeconf}  % 
\pdfoutput=1

\IEEEoverridecommandlockouts                              

\makeatletter
\let\NAT@parse\undefined
\makeatother
\usepackage[numbers]{natbib}

\usepackage{subcaption}
\usepackage{graphics}
\usepackage{amsmath}

\usepackage[colorinlistoftodos,prependcaption,textsize=small]{todonotes}
\usepackage{changes}

\title{\LARGE \bf
Learning Topological Motion Primitives for Knot Planning
}

\author{Mengyuan Yan$^{1}$ and Gen Li$^{2}$ and Yilin Zhu$^{1}$ and Jeannette Bohg$^{1}$% <-this % stops a space
\thanks{*The Okawa Foundation and Toyota Research Institute ("TRI") provided funds to assist the authors with their research but this article solely reflects the opinions and conclusions of its authors and not the Okawa Foundation, TRI or any other Toyota entity.}% <-this % stops a space
\thanks{$^{1}$Mengyuan Yan, Yilin Zhu and Jeannette Bohg are with School of Engineering, Stanford University, CA, USA, 94305
        {\tt\small \{mengyuan, ylzhu, bohg\}@stanford.edu}}%
\thanks{$^{2}$Gen Li is with Department of Computer Science and Technology, Tsinghua University, Beijing, China, 100084. This work was done while he was a research intern at Stanford University.
        {\tt\small lig16@mails.tsinghua.edu.cn}}%
}

\begin{document}

\maketitle
\thispagestyle{empty}
\pagestyle{empty}

\begin{abstract}
    In this paper, we approach the challenging problem of motion planning for knot tying. We propose a hierarchical approach in which the top layer produces a topological plan and the bottom layer translates this plan into continuous robot motion. The top layer decomposes a knotting task into sequences of abstract topological actions based on knot theory. The bottom layer translates each of these abstract actions into robot motion trajectories through learned topological motion primitives. To adapt each topological action to the specific rope geometry, the motion primitives take the observed rope configuration as input. We train the motion primitives by imitating human demonstrations and reinforcement learning in simulation. To generalize human demonstrations of simple knots into more complex knots, we observe similarities in the motion strategies of different topological actions and design the neural network structure to exploit such similarities. We demonstrate that our learned motion primitives can be used to efficiently generate motion plans for tying the overhand knot. The motion plan can then be executed on a real robot using visual tracking and Model Predictive Control. We also demonstrate that our learned motion primitives can be composed to tie a more complex pentagram-like knot despite being only trained on human demonstrations of simpler knots.
\end{abstract}

\vspace{-5pt}
\section{Introduction}
Autonomous manipulation of deformable objects has many potential applications, including robotic surgery, assistive dressing, textile and clothing manufacturing, etc. Tying knots with linear objects, e.g. ropes, is a common but challenging task in this research direction. Various types of knots are used in surgery or search and rescue. By teaching robots to accomplish these tasks, we could aid surgeons in saving more patients, and replace rescue task forces in high-risk environments. However, teaching robots to manipulate ropes and tie knots is hard. Ropes have a high-dimensional state space which makes visual perception challenging. They have a large action space and under-actuated dynamics, making common planning algorithms either non-applicable or computationally expensive. In addition, knot tying is a long-horizon task for which some local planners for deformable linear objects could fall into local minima~\cite{DLOplan2006,DLOplan}. 

\begin{figure}
    \centering
    \includegraphics[width=\columnwidth]{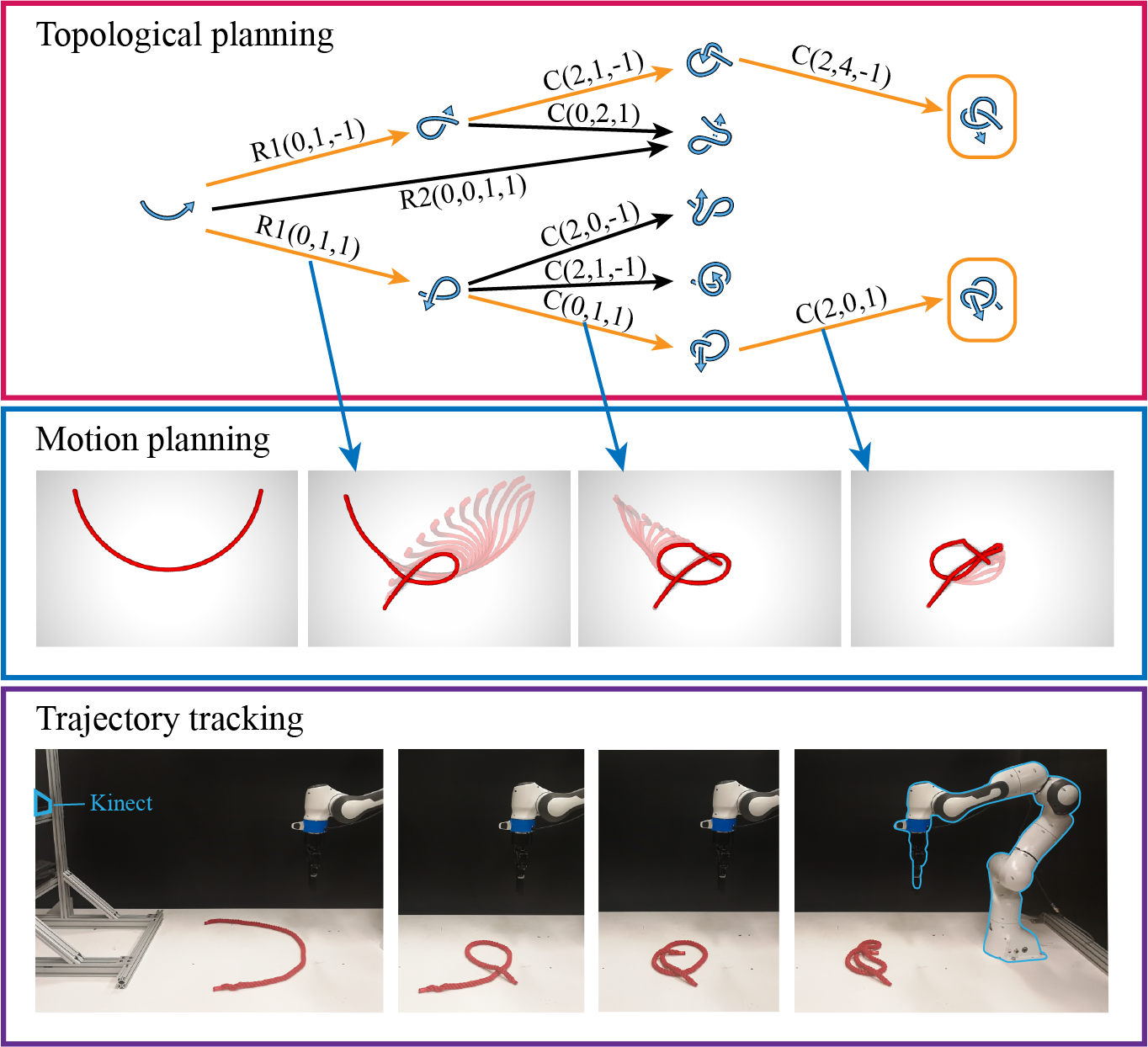}
    \caption{The 3-level inference process for tying a knot. On the top level, we search for topological paths from the start topology to the goal topology in the graph defined by knot theory. On the second level, we ground each edge in the topological path, which is a topological action, into robot motion trajectories using trained motion primitives. Each motion primitive will predict a robot motion spline curve conditioned on the geometric configuration of the rope at the end of the previous stage, and the predicted motion is simulated to obtain rope configurations during the manipulation. On the bottom level, the simulated rope configurations will be used as reference states for MPC to track on the real robot.}
    \label{fig:teaser}
\end{figure}

Knot theory and topology can help to guide motion planning for tying knots. For example, it can provide an abstract action skeleton, i.e. a sequence of topological actions, to decompose the long-horizon knotting task into shorter subtasks~\cite{Knotting2004}. It can also help to prune out unpromising edges in tree-based or roadmap-based motion planning methods~\cite{Knotting2008}. However, motion planning for each topological action remains challenging, and brute-force search for the physical robot motions is prohibitively expensive due to the high-dimensional action space and lack of intuitive cost functions to abstract topological goals. 

Learning from demonstrations provides a promising alternative. \citet{Surgical2010, Surgical2012} learn to clone and enhance the robot trajectories for tying surgical knots from human teleoperation. However, the learned robot motion does not generalize to new string configurations. We try to benefit from both knot theory and data-driven methods, and propose to learn a library of topological motion primitives which ground abstract topological actions in robot motions and rope geometries. The motion primitives can be composed to plan for various knots. We design the learning algorithm to facilitate learning from few human demonstrations and generalize to new rope configurations and new knotting tasks.

Our key contributions are:
\begin{itemize}
    \item We define \textit{topological motion primitives}, which translate abstract topological actions into concrete robot motion trajectories, conditioned on input rope geometric configurations. They can be composed to solve a variety of knotting tasks.
    \item We encode topological actions such that each topological motion primitive can instantiate a range of them. Human demonstrations are only needed for some topological actions and the topological motion primitives learn to generalize to unseen rope shapes and topological actions through reinforcement learning. 
    \item We demonstrate that the topological motion primitives can be composed to accomplish knot tying tasks on a real robot. We also demonstrate that we can plan for more complex knotting tasks than seen during training, which has not been shown in previous works.
\end{itemize}

%\vspace{-5pt}
\section{Related Work}
\paragraph{Planning for rope manipulation}
Planning for rope manipulation is a challenging task because of its high dimensional state space and highly under-actuated dynamics.
\citet{DLOplan2006,DLOplan} propose local planners for Deformable Linear Objects (DLOs) based on minimum energy curves. They can be used with tree-based and roadmap methods to search for longer plans. Although \cite{DLOplan2006, DLOplan} restrict the action space to the two ends of the rope, planning can take hours for a simple deformation task. 
For the long horizon and complex task of knotting, \citet{Knotting2008} build search trees by sampling robot motions, and use knot theory to prune unpromising branches. The high computational cost for search-based methods call for the use of human knowledge to bias the search space. \citet{Knotting2004} design strategies for choosing grasping points and robot motions for each topological action step in knotting tasks, but only verified their strategy in an untangling task. Our work proposes to learn such strategies from human demonstrations, which we call \textit{topological motion primitives}. By learning such primitives we generalize from very few human demonstrations to different geometrical and topological conditions, and greatly improve the efficiency of tree-based planning by intelligently biasing the sampling function.

\paragraph{Learning for rope manipulation}
Several recent works have learned dynamics models for ropes and used them with \textit{Model Predictive Control} (MPC) for manipulation tasks. \citet{PropNet} and \citet{InteractionNet} model ropes as mass-spring systems, and use graph networks to learn rope dynamics. \citet{DVF_journal} learn a video prediction model, without any physical concept of objects or dynamics. In follow-up works, \citet{Chelsea17, Chelsea18} investigate different image losses. \citet{DLO-RL} uses model-based reinforcement learning (RL) for deforming a rope in 2D with fixed start and goal configurations. While these works have demonstrated short-horizon deformation tasks, the method does not directly extend to knot planning, where the goal state is abstract, represented by topology. \citet{ZSVI} learn an inverse model that can predict robot actions for local deformation tasks, however humans are responsible for providing a visual plan that the inverse model will follow. Most related to our work, \citet{CausalInfoGan} tackles the long-horizon planning problem by embedding images into a plan-able latent space where neighboring points in the latent space correspond to images that are temporally close. Plans are generated using A* search in the latent space and decoded to observations. \citet{Surgical2010, Surgical2012} learn surgical knotting skills from human demonstrations, however only the robot end-effector trajectory is considered, and the state of the thread being manipulated is not included in the algorithm.  Our work is different from the above works in the following aspects. (1) We address the long-horizon problem of knot tying, while \cite{PropNet,InteractionNet,DVF_journal,DLO-RL} only addresses the problem of deforming a rope to a goal shape. (2) We build on the previous work~\cite{yan2019selfsupervised} which demonstrates estimation and tracking of rope states from images, so that planning and control can be done in state space instead of pixel space~\cite{DVF_journal,ZSVI,CausalInfoGan}. (3) We learn from human demonstrations and generalize to unseen situations by reinforcement learning, while previous works~\cite{PropNet,InteractionNet,DVF_journal,ZSVI,CausalInfoGan,DLO-RL} only use self-supervised robot exploration data and are unlikely to have seen successful multi-cross knots, and \cite{Surgical2010,Surgical2012} disregards the rope states and cannot generalize to new rope states. (4) Having access to the physical state of ropes, we use knot theory to decompose various knotting tasks into the same library of topological motion primitives, so that we can learn them with demonstrations on simpler knotting tasks, and compose the learned primitives to complete more complex knotting tasks.

%\subsection{learning for tree search papers?}

%\begin{itemize}

%\item \textbf{Learning Implicit Sampling Distributions for Motion Planning }
%Learning sample distribution in state space.

%\item \textbf{Efficient Sampling With Q-Learning to Guide Rapidly Exploring Random Trees}
%The general formulation of state value and Q-value is similar to ours, however the learning method and inference details are not great.

%\item \textbf{Guiding the search in continuous state-action spaces by learning an action sampling distribution from off-target samples }
%Interesting how they use importance sampling and large amounts of non-success samples to replace the small amount of success samples.

%\item \textbf{Demonstration-Guided Motion Planning }
%using demonstrations to learn promising state trajectory and time varying cost map (to model implicit task constraints).

%\end{itemize}
%\jean{same as earlier. Good summary of relevant papers. But what do you contribute? And how are you different? Or what are you making possible that wasn't possible before? Or how can you now do something faster or more accurate?}

%\section{Method}
\section{Problem Statement}
We define the task of robotic rope knotting as follows. The robot observes an untangled rope in its work space. The task is specified by a goal topological state, whose representation we will define in Sec. IV. The robot needs to find a plan of grasping points and motion trajectories that can bring the current rope configuration to a goal configuration with the desired goal topology, and execute the motion plan with visual tracking and MPC.

As shown in Fig.~\ref{fig:teaser}, we decompose the task of tying knots of user-specified topology into 3 levels. At the top level, knot theory defines a graph with topological states as nodes and topological actions as edges. Possible topological action sequences (topological plans) are obtained by graph search. At the second level, we translate each topological action into a motion trajectory of the robot and rope, conditioned on the observed configuration of the rope. We call the mapping functions {\em topological motion primitives\/}. Once the preceding topological action has been translated into a robot motion trajectory, the rope trajectory is obtained via simulation, and the end configuration serves as the start configuration for the subsequent topological action.
At the bottom level, we execute the motion plan on the real robot. We use a visual tracking algorithm together with MPC to track the planned rope trajectory in a closed-loop manner, adjusting the robot motion commands to account for errors in rope dynamics and for noise in perception and actuation. 

\section{Background on topology}

\begin{figure}
    \centering
    \includegraphics[width=0.35\textwidth]{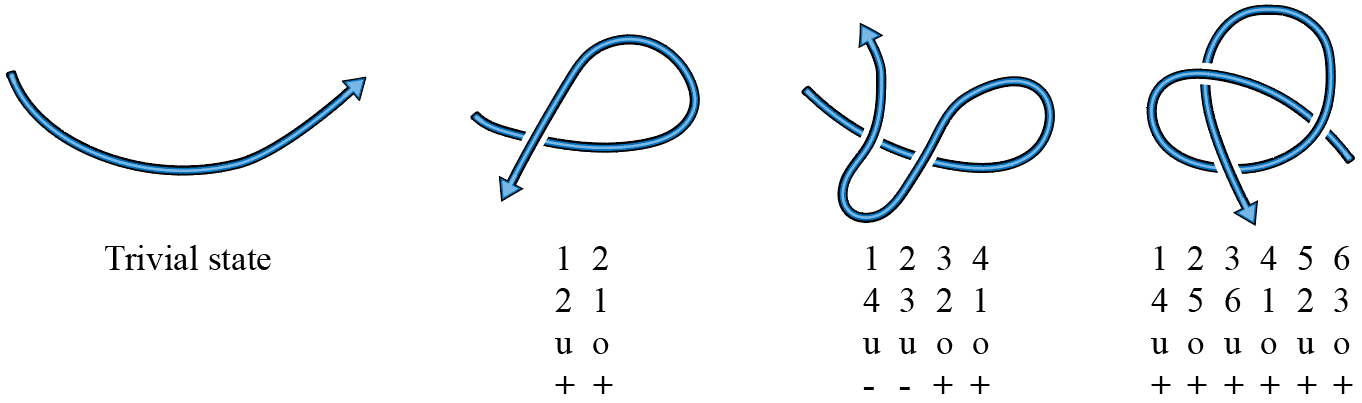}
    \caption{Example geometric configurations and the representations of corresponding topological states. The arrow represents the direction from head to tail. "o"/"u" indicate the vertical order of each intersection (short for over/under), and "+"/"-" indicate the sign of each intersection.}
    \label{fig:topology-state}
\end{figure}

\vspace{-3pt}

\begin{table}[]
    \centering
    \begin{tabular}{||c||c||}
    \hline  & \\[-0.9em]  \hline
        Reidemeister I (R1) &  Reidemeister II (R2) \\
        \hline
        \includegraphics[width=0.2\textwidth]{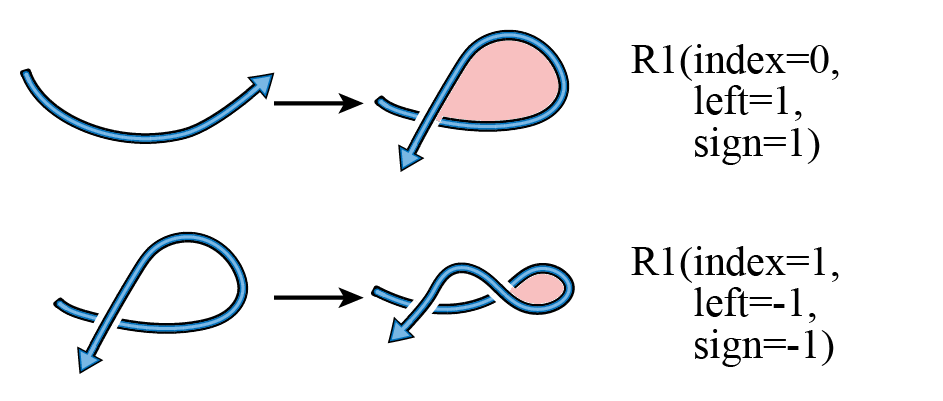} & 
        \includegraphics[width=0.2\textwidth]{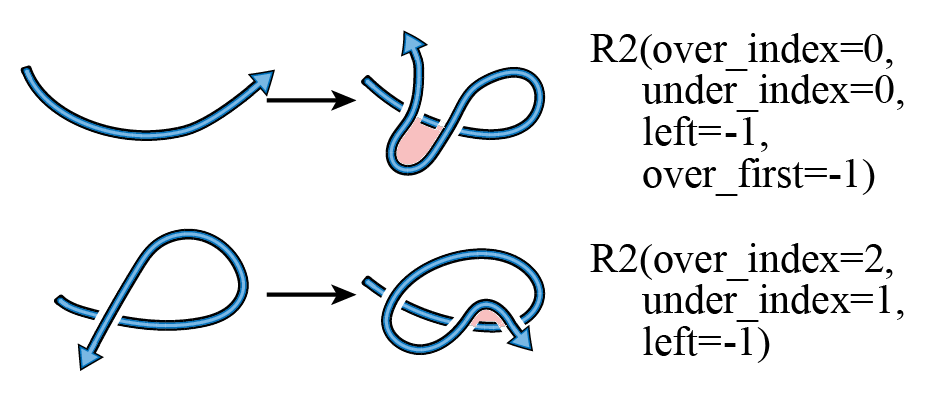} \\
        \hline  & \\[-0.9em] \hline
        Cross (C) & Reidemeister III (R3) \\
        \hline
        \includegraphics[width=0.2\textwidth]{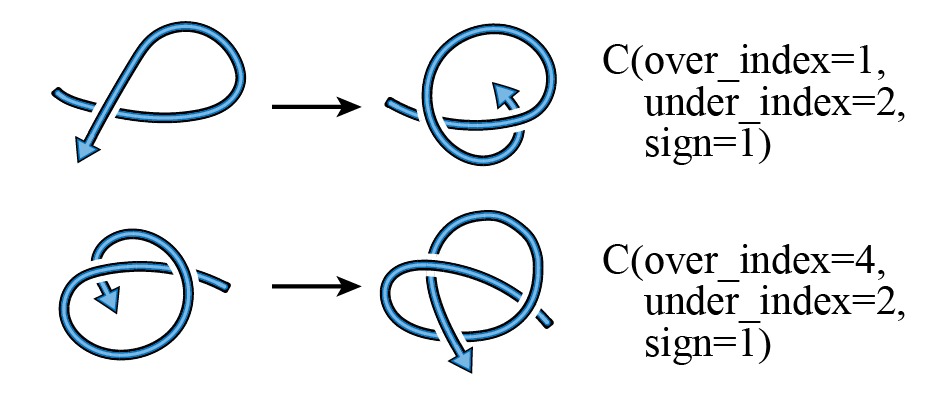} &
        \includegraphics[width=0.2\textwidth]{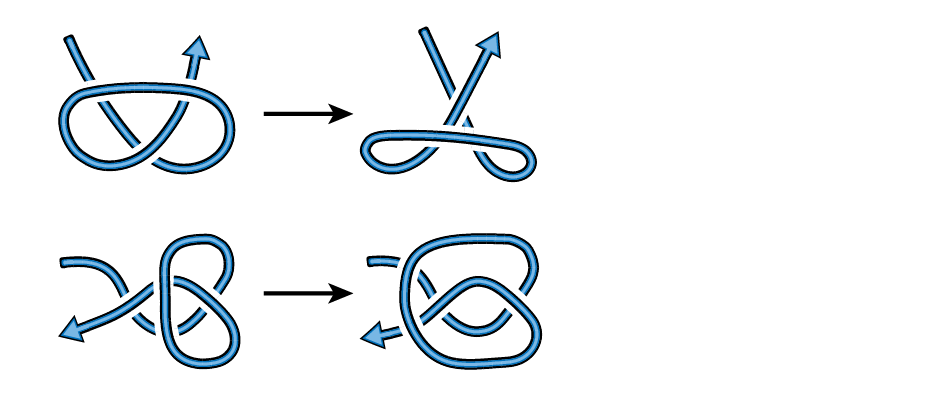} \\
        \hline 
    \end{tabular}
    
    \begin{tabular}{||p{1.37cm}|p{5.8cm}||}
    \hline \hline
        index &  For R1 only. Index of the segment forming the new intersection\\
        \hline
        over\_index & For R2 and C. Index of the segment that will be over the new intersection(s). \\
        \hline
        under\_index & For R2 and C. Index of the segment that will be under the new intersection(s). \\
        \hline
        left & For R1 and R2. Binary. Whether the new face (shaded) is on the left side of the top segment. \\
        \hline
        sign & For R1 and C. Binary. Sign of the new intersection. \\
        \hline
        over\_first & For R2 if over\_index = under\_index. Binary. Whether the half segment closer to head will be over the new intersections.\\
        \hline
    \end{tabular}
    \caption{Top: examples actions in each category and their parameter values. Bottom: Definitions of the parameters.}
    \label{tab:topology-action}
\end{table}
We follow the representation of a rope's topological state in \cite{Knotting2003}. A topological state is defined based on a 2D projection of the 3D curve. We use the horizontal plane in the world frame as the projection plane. Some examples of projected curves and their topological state representation are shown in Fig.~\ref{fig:topology-state}. Given a projection, we pick one end of the rope as the head and the other end as the tail. As we trace the curve from head to tail, we number the intersections starting from $1$. Each intersection will be encountered twice and therefore receives two numbers. Then, we retrace the curve, and when we encounter an intersection, record the two numbers from the current and the other intersecting strand, as well as a sign plus/minus and a relative vertical position (over/under) between the current and the other intersecting strand. A sign is determined by the following equation:
\begin{equation}
    sign = \frac{\vec{l}_{over} \times \vec{l}_{under}}
                {|\vec{l}_{over} \times \vec{l}_{under}|} \cdot \vec{e_z},
\end{equation}
where $l_{over}$ and $l_{under}$ are the directional vectors for the two strands, and $e_z$ is the unit normal of the projection plane.

To transition between the topological states, there are four categories of topological actions, and each category has many topological action instances, indexed by a few discrete parameters. Examples for each category are shown in Table~\ref{tab:topology-action} (top) and the definition of parameters are listed in Table~\ref{tab:topology-action} (bottom). Segments of the projected curve, separated by intersection points, are indexed starting from 0 when tracing the curve from head to tail. 
\begin{itemize}
    \item The Cross (C) action makes a new intersection using the head/tail segment with another segment. 
    \item The Reidemeister I (R1) action makes a new loop using one segment of the curve. 
    \item The Reidemeister II (R2) action makes two new intersections of opposite signs, by pulling the middle of one segment on top of another segment.
    \item The Reidemeister III (R3) action moves two neighboring intersections to the other side of a cross. We do not consider this more complex action category in our planning.
\end{itemize}

Note that these topological action categories are different from the grouping into topological motion primitives in Sec.~\ref{sec:meta_prim}.

All topological states and actions form a directed acyclic graph (DAG), with the trivial topological state (i.e. untangled) as the root. Given a goal state, we can find one or more paths from the trivial topological state. Fig.~\ref{fig:teaser} (top) shows a partial graph for the example of an overhand knot. The overhand knot has two topologies with opposite chirality. We show 2 of the 8 possible different topological paths.

\vspace{-5pt}
\section{learning topological motion primitives}
Now we want to translate the abstract topological actions obtained from graph search into concrete robot motion trajectories and rope motion. The downstream closed-loop controller will use the resulting rope and robot trajectories as a reference. 

We made two design choices in learning the topological motion primitives. First, the action space of the motion primitives will be long robot motion trajectories parameterized by spline curves, instead of small delta positions commonly used for feedback policies. This choice drastically shortens the horizon of the resulting RL problem so that the training process will be stable and more data-efficient, even with simple RL algorithms. Meanwhile, spline curves are complex enough to allow successful knotting with only a few re-grasps. Second, each topological motion primitive needs to work for a group of topological actions in order to scale to more complex knotting tasks. Thus we encode the discrete topological action parameters, listed in Table~\ref{tab:topology-action}, as an input to the motion primitives. The encoding is designed to facilitate learning of the similarities among motion policies that humans demonstrate for different topological actions.

We want the topological motion primitives to learn distributions of successful motion splines, conditioned on the current rope shape and the topological action to instantiate. Samples can be drawn from the learned distribution and used in tree-search during inference. Each spline curve has 3 control points. The first control point will be on the rope and serves as the grasping point. The last control point will be on the supporting table plane so that the robot gently releases the rope instead of dropping. The middle point can be above the table plane and determines the maximum height of the spline. This formulation leads to a 6D action space. We observe that in most cases one motion spline is enough to accomplish a topological action. For the cases where re-grasping is necessary, we expect that sequencing a few such spline curves could solve the problem. However, we leave that to future work.

\subsection{Two types of similarities in motion policies}
\label{sec:meta_prim}
In this section, we will introduce two types of similarities in motion policies for different topological actions, to help the topological motion primitives generalize. Based on these similarities, we group all topological actions in the R1, R2 and Cross category into 4 topological motion primitives:
\begin{description}
    \item[Set 1:] All R1 actions.
    \item[Set 2:] All R2 actions.
    \item[Set 3:] Cross actions where the top segment is one end.
    \item[Set 4:] Other Cross actions.
\end{description}
Cross actions are split into 2 sets based on the similarities in grasping position on the rope.

\begin{figure}
    \centering
    \includegraphics[width=0.45\textwidth]{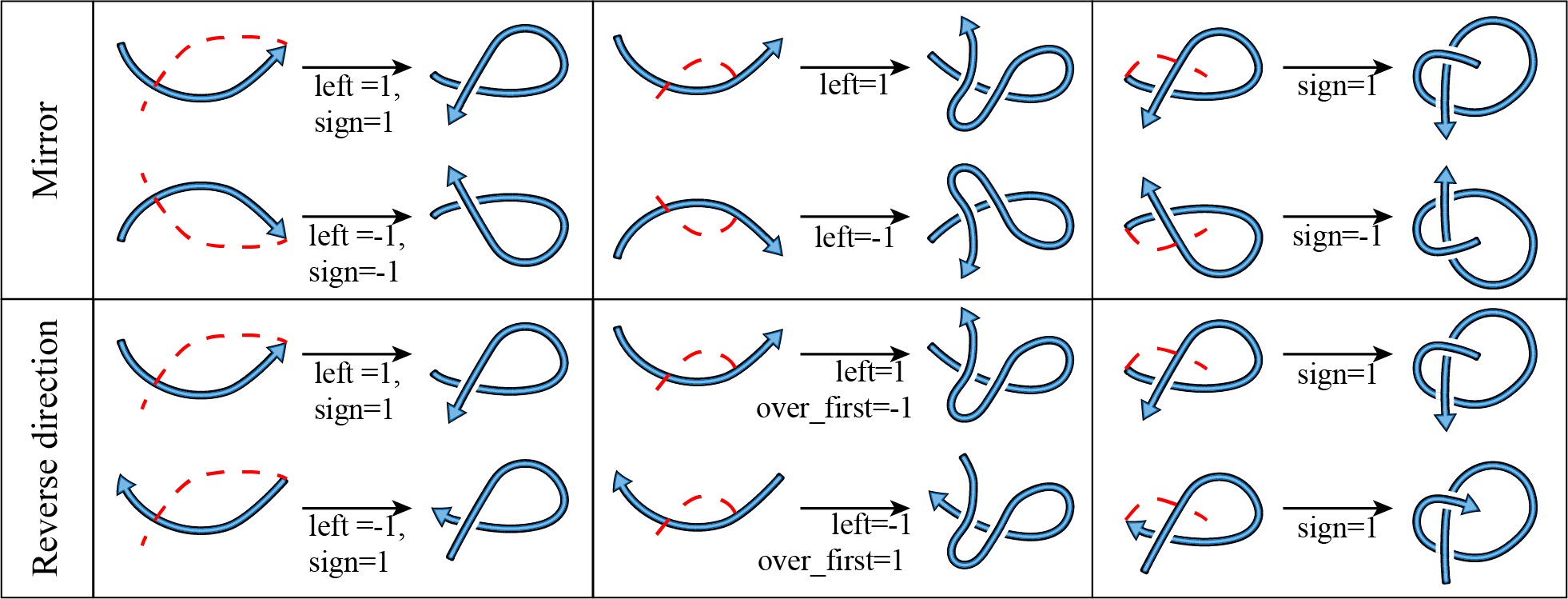}
    \caption{Examples applying the \textit{Mirror} or \textit{Reverse} transformation on the geometric configurations and predicted robot motion splines. From Left to Right: Examples for the R1, R2 and Cross category. Applying mirror transformation to rope shapes and robot trajectories will change the corresponding topological action to have the opposite "sign" and "left" parameters. Reversing the rope geometries will change the corresponding topological action to have the opposite "left" and "over\_first" parameters.}
    \label{fig:merge}
\end{figure}

The first type of similarity is based on spatial symmetries in state transitions. The symmetries allow to reduce the number of topological actions that need to be learned. The topological actions that are not directly learned can be instantiated by using their learned counterparts, and applying geometric transformations to the rope configurations and predicted splines. 
Fig.~\ref{fig:merge} shows examples for the two transformations: \textit{Mirror} and \textit{Reverse}. Each cell in the figure shows a pair of transitions $(s, a, s')$ before (top in each cell) and after (bottom in each cell) applying the indicated transformation. Here $s$ and $s'$ refer to the geometric rope configuration and $a$ is the  robot motion spline shown as red dashed curves. Each transition is also described, in the topological level, as a transition $(s_\textrm{topo}, a_\textrm{topo}, s_\textrm{topo}')$, and binary-valued parameters for the topological action $a_\textrm{topo}$ are shown below the arrows.
The \textit{Mirror} transformation reflects the rope geometries and the robot trajectory about the horizontal axis (in fact, reflecting about any line in the plane has the same effect, but we choose the horizontal axis in our implementation). The corresponding topological actions before and after this transformation will have opposite "left" and "sign" parameters, but the other parameters will be the same. The \textit{Reverse} transformation switches the head and tail of the rope, but keeps the robot trajectory unchanged. The corresponding topological actions before and after this transformation will have opposite "left" and "over\_first" parameters, segment indices will also change. Either transformation, when applied twice, will be identity and produce no change.

As a result, it is sufficient to learn, for example, the topological action R1(index=0, left=1, sign=1), and we can use the same network to infer robot motion splines for R1(index=0, left=$\pm$1, sign=$\pm$1). For example, to infer the motion spline for R1(index=0, left=-1, sign=-1), we mirror the start state, feed the transformed state into the network to predict spline parameters, and then apply the inverse transformation, i.e. mirroring, to the predicted spline parameters. To infer the motion spline for R1(0, -1, 1), we use the \textit{Reverse} transform instead, and for R1(0, 1, -1) we use the combination of both transformations.

\begin{figure}
    \centering
    \includegraphics[width=0.47\textwidth]{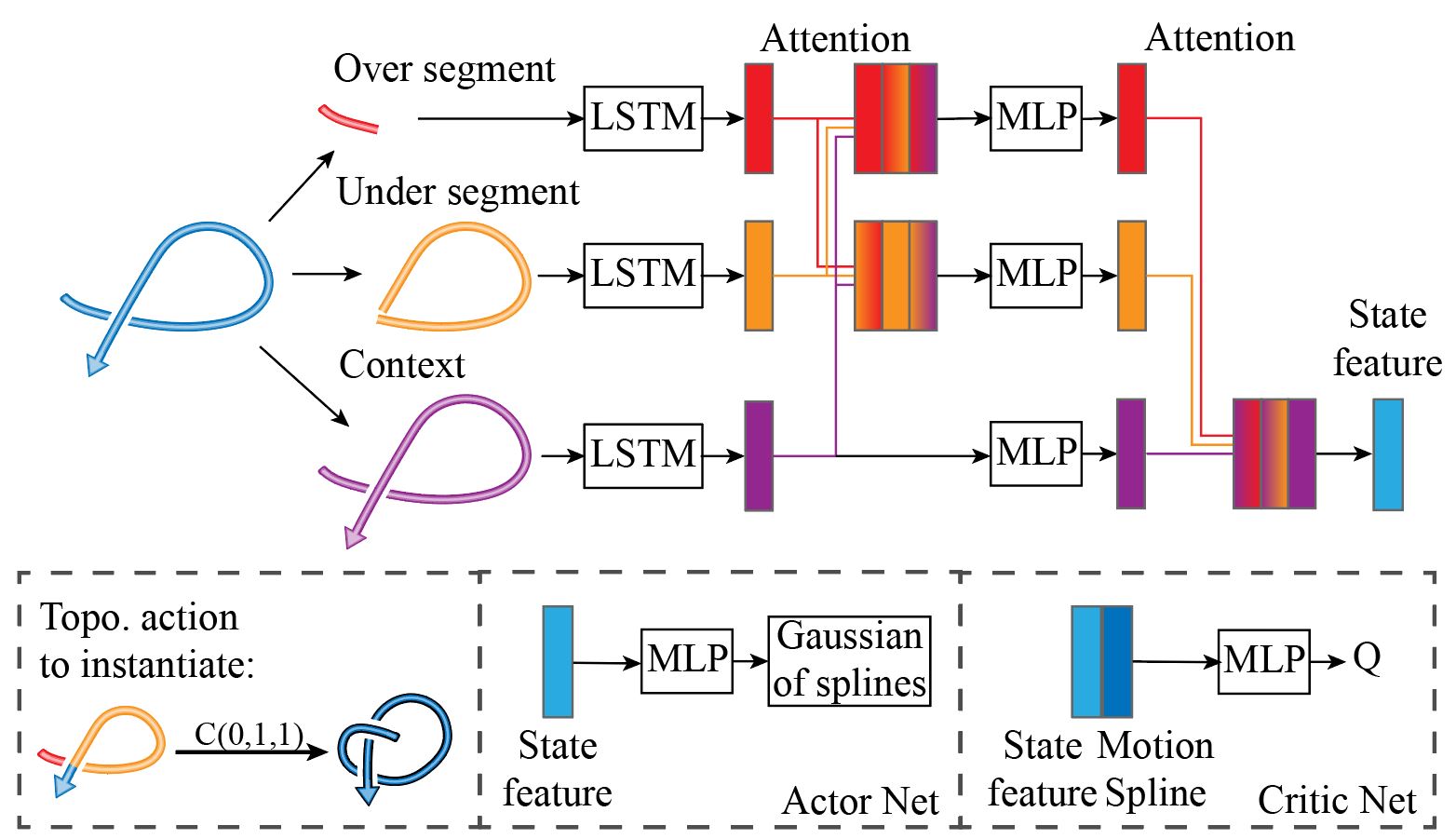}
    \caption{Illustration of input encoding and network structure for topological motion primitives. Top: The trunk network for both the actor and critic network. The rope's geometric configuration is encoded in 3 parts according to the topological action to be instantiated, which is illustrated on the bottom left. The rectangles with color gradients represent the output features of attention modules~\cite{attention}, the color towards the left (right) edge of the rectangle is the color of the query (value) feature to the attention module respectively. The LSTM and MLP for each stream do not share weights. Bottom middle and right: prediction for actor network and critic network, based on the feature extracted from the trunk network.}
    \label{fig:network}
\end{figure}

The second type of similarity is more conceptual. When making a new intersection, humans will grasp the segment that will be over the newly formed intersection, and move towards the segment that will be underneath. This common strategy is captured by encoding the rope's geometric configuration into three parts: The segment which will be over the newly formed intersection, the segment which will be underneath, and the whole state which acts as task context to e.g. avoid undesired self-collision. For R1 actions, the over and under segment will be the same. The network structure to process this input is shown in Fig.~\ref{fig:network}. Pairwise attention modules~\cite{attention} are used to propagate information between the three streams.

\subsection{Training and inference process}\label{sec:training_prim}
We will train the topological motion primitives in two stages: (i) reinforcement learning for single-step tasks, where each motion primitive is trained independently and demonstration data is used to bias initial exploration, and (ii) reinforcement learning for multi-step knotting tasks, where the motion primitives are sequenced to form the policy and are trained jointly.

\paragraph{Collecting demonstration} With our action space formulation, demonstration is simply given by clicking on three points on an image of the start geometric configuration, indicating the position of the three control points of the spline. Gaussian noise is added to the annotation to generate sample motion splines, and the motions are simulated to verify if the desired topological action is accomplished. All the trials are saved and we refer to them as demonstration data.

\begin{figure}
    \centering
    \includegraphics[width=0.37\textwidth]{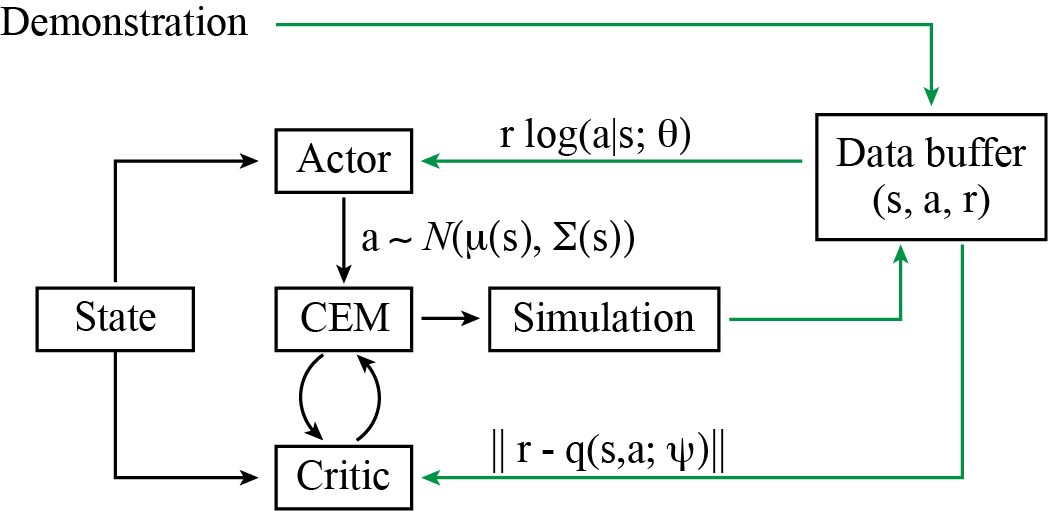}
    \caption{Illustration of the training and inference process for motion primitives. States $s$ are geometric configurations of the rope. Actions $a$ are 6D vectors parameterizing spline curves. Rewards $r$ are binary, indicating success of the desired topological action. The critic network with weights $\psi$ predicts Q values and the actor network with weights $\theta$ predicts the mean and variance $\mu, \Sigma$ for Gaussian action distributions. The black arrows constitutes the policy during inference, as well as the behavior policy during training. Green arrows indicate data flow and objectives during training.}
    \label{fig:training}
\end{figure}

\paragraph{Independent training on single-step tasks} We adapt QT-Opt~\cite{QTOpt} to train each topological motion primitive. The training process is summarized in Fig.~\ref{fig:training}. QT-Opt trains a critic neural network to approximate the Q function, and uses the \textit{Cross-Entropy Method} (CEM) to find the action with maximum Q value as the policy. However, we find this to be in-sufficient for our high dimensional action space. Thus, we also train an actor network to predict Gaussian distributions conditioned on the start configuration of the rope to initialize the CEM process. The actor network is trained with REINFORCE~\cite{REINFORCE} with data from the replay buffer. Although other methods such as Trust Region Policy Optimization or Proximal Policy Optimization can be used, we expect them to bring little benefit, since with our formulation of the action space, the resulting RL problem has a short horizon. The replay buffer is initially populated with demonstration data, and this training stage will be referred to as imitation learning. New data are added to the replay buffer once imitation learning has converged, and this training stage will be referred to as reinforcement learning.

\paragraph{Joint training on multi-step tasks}
There are infinitely many geometric configurations that correctly instantiate a topology. Some configurations are more favorable than others because they make the next action easier. It is hard to engineer a continuous reward function which can bias the motion primitives towards generating more favorable end states. 
Luckily, such a reward function is available once we trained the library of motion primitives. Let us say that the topological path found for a task is $(s_{\textrm{topo}}, a_{\textrm{topo}}, s_{\textrm{topo}}^\prime, a_{\textrm{topo}}^\prime, s_{\textrm{topo}}^{\prime\prime})$, and geometric instantiations of this topological path is $(s,a,s^\prime, a^\prime, s^{\prime\prime})$. The state value function $V_{a_{\textrm{topo}}^\prime}(s^\prime) = \max_{a^\prime} Q_{a_{\textrm{topo}}^\prime}(s^\prime, a^\prime)$ is the predicted success rate of $a_{\textrm{topo}}^\prime$ starting from configuration $s^\prime$. This value thus is a reward for $a_{\textrm{topo}}$, $r_{a_{\textrm{topo}}}(s,a)$.  
Based on this observation, we finetune the topological motion primitives jointly on multi-step knotting tasks, where the RL horizon is set to the number of topological actions in the path found during topological graph search. The policy only receives a reward of 1 at the end of the episodes if the desired knot topology is achieved. Different from a normal RL problem, different motion primitives are used for each time step in an episode according to the topological plan, and each motion primitive receives different transitions from the replay buffer as training data. The same adapted QT-Opt algorithm is used for this training process.

\paragraph{Grounding the topological path: tree search} Once the four topological motion primitives are trained, we use them as action samplers in tree search, to guarantee finding a motion plan. From the top level topological planning, we have found one or more topological paths from the untangled state to the given knot topology. The paths form a DAG, which we call the solution DAG. We start from the geometric configuration of the untangled state as the tree root. At each tree expansion step, we randomly select a node in the tree (including both leaf nodes and non-leaf ones), biasing the selection probability towards nodes whose topological states are closer to the knot topology in the solution DAG (measured by the number of edges on the shortest path in the solution DAG). From the selected node, we randomly select an outgoing edge in the solution DAG as the next topological action to instantiate. We use the corresponding topological motion primitive to predict a motion spline for the robot, and this spline is executed in simulation to obtain the rope motion trajectory. Note that the predicted spline is not guaranteed to reach the desired rope topology prescribed by a topological action. Therefore, the rope's end configuration together with its topology is only added to the search tree if the goal topology of the topological action is reached. The tree keeps expanding until the final knot topology is reached.

\vspace{-3pt}
\section{closed-loop control and visual tracking}
To execute the planned robot motion on the real robot, we sample way points from the robot motion spline, and record corresponding rope geometric configurations from the simulator. We use the rope motion trajectories as reference states for MPPI~\cite{MPPI} that tracks this rope motion. The planned robot motion spline is used to initialize the actions used in MPPI. We use the perception network in \cite{yan2019selfsupervised} to estimate the initial rope state, and use the image space loss and LSTM dynamics model in \cite{yan2019selfsupervised} to track the 3D rope state across time. We extended the differentiable rendering of rope states to 3D, by additionally taking the Kinect depth images.

\vspace{-5pt}
\section{Experiments}
We evaluate four aspects of our proposed method. First, we quantitatively verify two hypotheses underlying our method: using human demonstrations to bias the search for robot motion splines can drastically reduce the number of samples compared to brute-force search, and using reinforcement learning on a wider range of start configurations improves generalization compared to only using imitation learning. Second, we show ablation studies that validate two design choices: using both the actor and critic networks for learning the motion primitives, and finetuning the motion primitives jointly on multi-stage knotting tasks. Both decisions improve success rates of the motion primitives and reduce search costs. Third, we compare our method to Causal InfoGAN~\cite{CausalInfoGan} as a baseline on knot planning. Finally, we qualitatively demonstrate that our generated plan can be executed on a real robot, and our method can plan for a knot more complex than seen during training.

\subsection{Value of using demonstrations and learning}
\label{sec:experiment1}

\subsubsection{Data}
We evaluate on eight topological actions, listed in Fig.~\ref{fig:stage-wise-baselines}. They are visualized in Fig.~\ref{fig:teaser} (Top). 
For R1(0,1,1) and R2(0,0,1,1), only one start configuration is demonstrated, which is a straight line. For each demonstration, 480 splines are generated by adding Gaussian noise, and simulated. For each of the other six topological actions, we demonstrate on eight start configurations and simulate 320 trials from each configuration. All the trials are saved as demonstration data. The eight topological actions are categorized into the four topological motion primitives as described in Sec.~\ref{sec:meta_prim}. Thus, the demonstration data is also separated into four buffers to train four pairs (actor and critic) of networks. The categorization is reflected by vertical lines in Fig.~\ref{fig:stage-wise-baselines}.

\subsubsection{Baseline}
Our first baseline is to randomly sample a motion spline from the whole parameter space. Our second and third baselines use the demonstration data but no further reinforcement learning. For the second baseline, generalization is achieved by interpolating demonstrations. For each demonstrated configuration, we fit a Gaussian distribution to all successful spline parameters. For new configurations, the {\em Iterative Closest Point} (ICP) distance to each demonstrated configuration is calculated. we use the exponential of negative ICP distances as the weights to interpolate the Gaussian parameters. Splines are sampled from the interpolated Gaussian distributions to evaluate the success rate. For third baseline, we train our topological motion primitives with imitation learning only.

\subsubsection{Single-stage evaluations}
\begin{figure}
    \centering
    \includegraphics[width=0.45\textwidth]{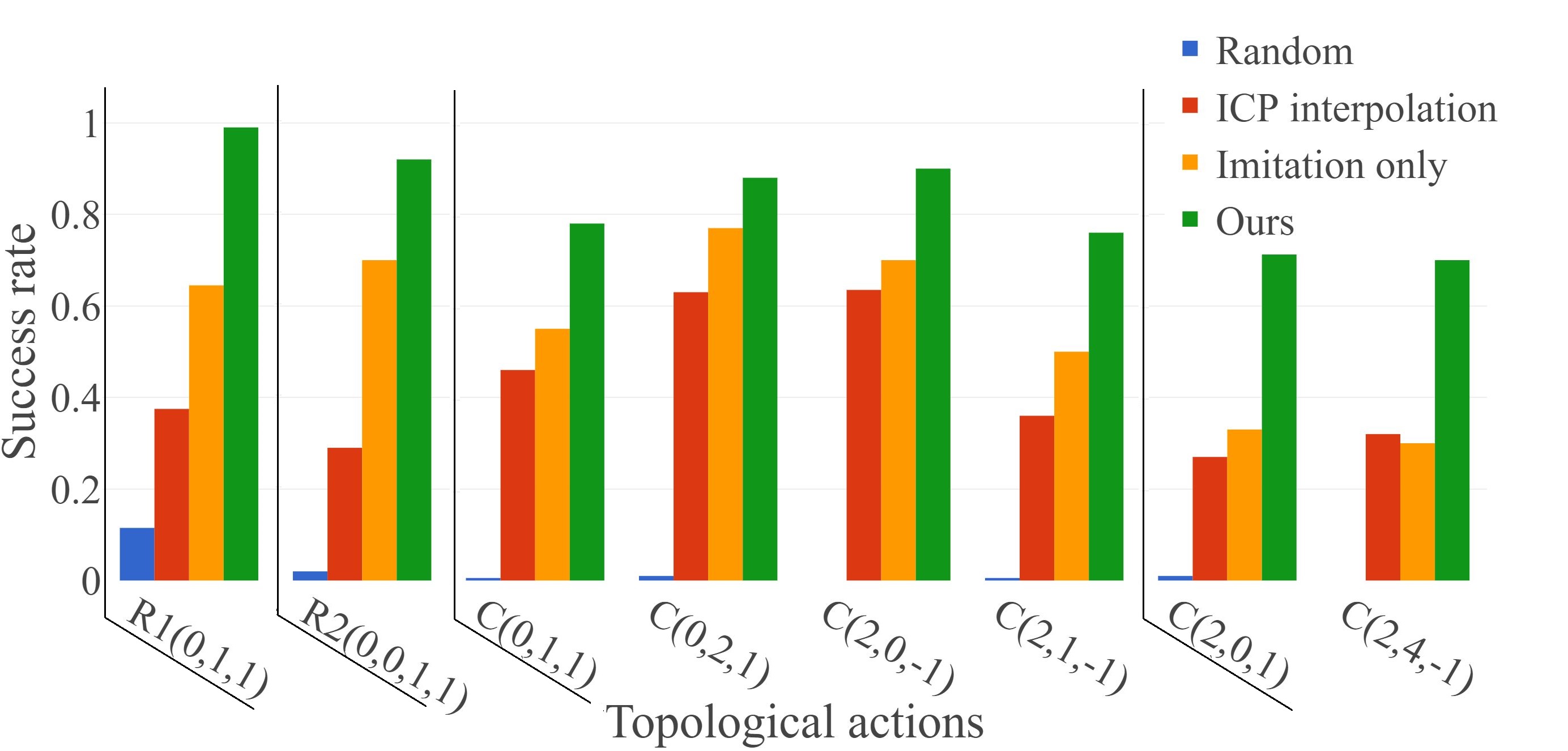}
    \caption{Success rate of predicted spline samples for eight topological actions, using a random sampler, the ICP interpolation of human demonstrations, our motion primitive networks only trained with imitation learning, and our motion primitives trained with imitation and reinforcement learning (Ours). Black vertical lines indicate categorization of topological actions into the four motion primitives.}
    \label{fig:stage-wise-baselines}
\end{figure}

For each topological action we evaluate the success rate of predicted spline samples for an unseen set of geometric configurations, but of the same topology as seen during training. The results are shown in Fig.~\ref{fig:stage-wise-baselines}. Comparing imitation-only motion primitives to brute force search, there are up to 100x improvements in success rates, thus reduction in search time, by using human demonstrations. The imitation-only motion primitives also have higher success rates than ICP interpolation on most topological actions, especially R1(0,1,1) and R2(0,0,1,1), suggesting the advantage of our neural network compared to hand-designed methods. By further training the motion primitives with reinforcement learning on a wider range of rope shapes, we improve the success rate by 1.14x to 2.33x compared to imitation-only.

% \begin{table}[]
%     \centering
%     \caption{Success rate of predicted spline samples for 8 different topological actions, using a random sampler, the ICP interpolation of human demonstrations, and our trained motion primitives.}
%     \begin{tabular}{||c|c|c|c||}
%     \hline
%     Topo. Action & Random & ICP interpolation & IL & Ours \\
%     \hline
%     R1(0,1,1)    & 11.5\% & 37.5\% & 64.5\% & 99\% \\
%     \hline
%     R2(0,0,1,1)  & 2\%  & 29\% & 70\% & 92\% \\
%     \hline
%     C(0,1,1)     & 0.5\% & 46\% & 55\% & 78\% \\
%     C(0,2,1)     & 1\% & 63\% & 77\% & 88\% \\
%     C(2,0,-1)     & 0\% & 63.5\% & 70\% &  90\% \\
%     C(2,1,-1)     & 0.5\% & 36\% & 50\% & 76\% \\
%     \hline
%     C(2,0,1)     & 1\% & 27\% & 33\% & 73\% \\
%     C(2,4,-1)    & 0\% & 32\% & 30\% & 70\% \\
%     \hline
%     \end{tabular}
%    \label{table:stage-wise-baselines}
% \end{table}

\subsubsection{Multi-stage knotting evaluations}
We also evaluate the number of branches in the search tree when planning for an overhand knot. For simplicity we only follow one topological path instead of all possible paths. This path involves a sequence of 3 topological actions, R1(0,1,1), C(0,1,1) and C(2,0,1). We use two start configurations and repeat 10 experiments for each configuration, with different random seed: (i) the start configuration is a straight line, which has been demonstrated when training the motion primitives, and (ii) the start configuration is an arc curve and generalization is necessary. When using the ICP interpolation, the tree size shows significant variance. Although the smallest trees only have 3 branches, which is the theoretical lower limit, the largest tree has 190 branches for the straight configuration and 1428 branches for the arc. When using the imitation-only motion primitives, the largest tree has 43 branches for the straight configuration and 9 branches for the arc. When using our motion primitives trained with both imitation learning and RL, the tree size is kept below 5 for all 10 seeds, and reaches the lower limit of 3 branches more than half of the time. This greatly reduced search time demonstrated our method's superior generalization to new geometries.

%\begin{table}[]
%    \centering
%    \caption{The number of branches evaluated in tree search from start configurations to an overhand knot.}
%    \begin{tabular}{||c|c|c|c|c||}
%        \hline
%        Start config. & Method & Min & Median & Max \\
%        \hline
%        Straight & ICP heuristic & 3 & 7.5 & 190 \\
%        Straight & Motion primitives & 3 & 3 & 4 \\
%        \hline
%        Curved & ICP heuristic & 3 & 14 & 1428 \\
%        Curved & Motion primitives & 3 & 3 & 5 \\
%        \hline
%    \end{tabular}
%    \label{table:knot-baselines}
%\end{table}

\subsection{Ablation studies}
In this section we evaluate two design choices: (1) Using both actor and critic networks for learning topological motion primitives, as compared to only using one of the network, and (2) finetuning the topological motion primitives jointly on the overhand knotting task.

\subsubsection{Benefit of using actor-critic}
\begin{figure}
    \centering
    \includegraphics[width=0.4\textwidth]{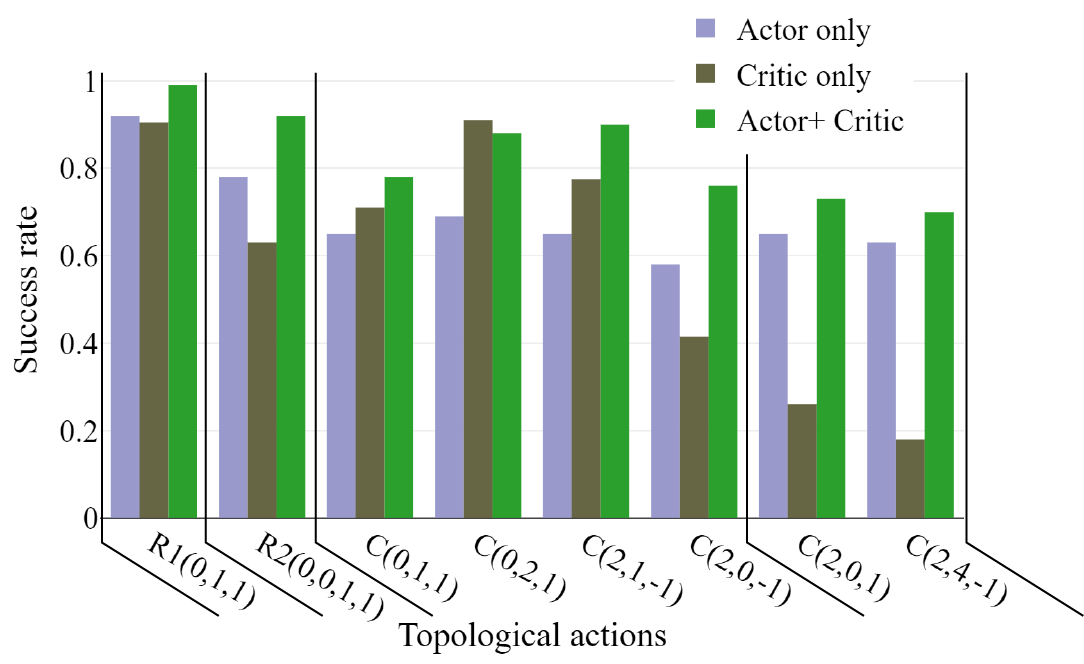}
    \caption{Success rate of predicted spline samples for 8 topological actions, using topological motion primitives trained using actor network only, critic network only, and both networks together.}
    \label{fig:stage-wise-ablation}
\end{figure}

We evaluate the success rate on the same sets of topological actions and rope configurations as used in Sec.~\ref{sec:experiment1}, and report the results in Fig.~\ref{fig:stage-wise-ablation}. When training the critic only, CEM is initialized with a large Gaussian covering the whole space of spline parameters, and is run for 10 iterations to find the best action. When training the actor only, we take samples from the predicted Gaussian distribution. When training both networks, the actor prediction is used to initialize the CEM, and CEM is run for only 1 iteration.

From Fig~\ref{fig:stage-wise-ablation}, training with both actor and critic networks achieves the best results for 7 out of 8 topological actions. The advantage of training both networks, over using the critic only, confirms that CEM is not effective enough in finding maxima in a high dimensional action space, which is also noted in \cite{Yan_2019}. The difference in success rate is particularly big for topological actions C(2,0,1) and C(2,4,-1), which require high accuracy robot motion. Using both actor and critic also performs better than training the actor alone, possibly for two reasons: (1) Gaussian distributions are not expressive enough to approximate the successful spline parameter distribution. (2) REINFORCE is an on-policy RL algorithm, but we are using it in our off-policy setting with replay buffers, because we can only afford a small amount of interaction with simulation. Therefore the actor is overly optimistic near the boundary of successful and failing spline parameters. We have experimented with correcting this bias with importance sampling, but could not stabilize the training. 

% \begin{table}[]
%     \centering
%     \caption{Success rate of predicted spline samples for 8 different topological actions, using topological motion primitives trained using actor network only, critic network only, and both networks together.}
%     \begin{tabular}{||c|c|c|c||}
%     \hline
%     Topo. Action & Actor only & Critic only & Actor + Critic \\
%     \hline
%     R1(0,1,1)   & 92\% & 90.5\% & \textbf{99\%} \\
%     \hline
%     R2(0,0,1,1) & 78\% & 63\%   & \textbf{92\%} \\
%    \hline
%     C(0,1,1)    & 65\% & 71\%   & \textbf{78\%} \\
%     C(0,2,1)    & 69\% & \textbf{91\%}   & 88\% \\
%     C(2,0,-1)   & 65\% & 77.5\% & \textbf{90\%} \\
%     C(2,1,-1)   & 58\% & 41.5\% & \textbf{76\%} \\
%     \hline
%     C(2,0,1)    & 65\% & 26\%   & \textbf{73\%} \\
%     C(2,4,-1)   & 63\% & 18\%   & \textbf{70\%} \\
%     \hline
%     \end{tabular}
%     \label{table:stage-wise-ablation}
% \end{table}

\subsubsection{Benefit of joint finetuning}
We repeat the planning process from randomly generated untangled start configurations to an overhand knot and evaluated the number of evaluated branches. When using the motion primitives trained independently, the number of evaluated branches is $4.3\pm 2.6$ over 50 experiments. After finetuning, the number drops to $3.4\pm 0.6$, showing notable improvement, especially for the worst cases.

\subsection{Comparison to Causal InfoGAN}
We compare our method to using Causal InfoGAN~\cite{CausalInfoGAN-workshop} on the same tasks of planning for single-stage topological actions and the multi-stage overhand knot.
We modified open source code~\cite{CausalInfoGAN-workshop} to use the configurations of the rope instead of images as observations. Causal InfoGAN learns to embed the rope configurations into a latent space where a linear-Gaussian stochastic transition model is prescribed. When planning a path between given start/goal configurations, the algorithm first project the given configurations into the latent space, i.e. search for a point in the latent space that maps to a configuration closest to the given start/goal, then interpolate the straight line in the latent space from the projected start point to the projected goal. The interpolation points are mapped to rope configurations as way points that constitute a plan. We trained the network using the demonstration data for only one topological action, C(0,1,1). We also tried to train the network on all demonstration data but cannot get visually plausible results, due to commonly known training instabilities of GANs.

\begin{figure}
    \centering
    \includegraphics[width=0.35\textwidth]{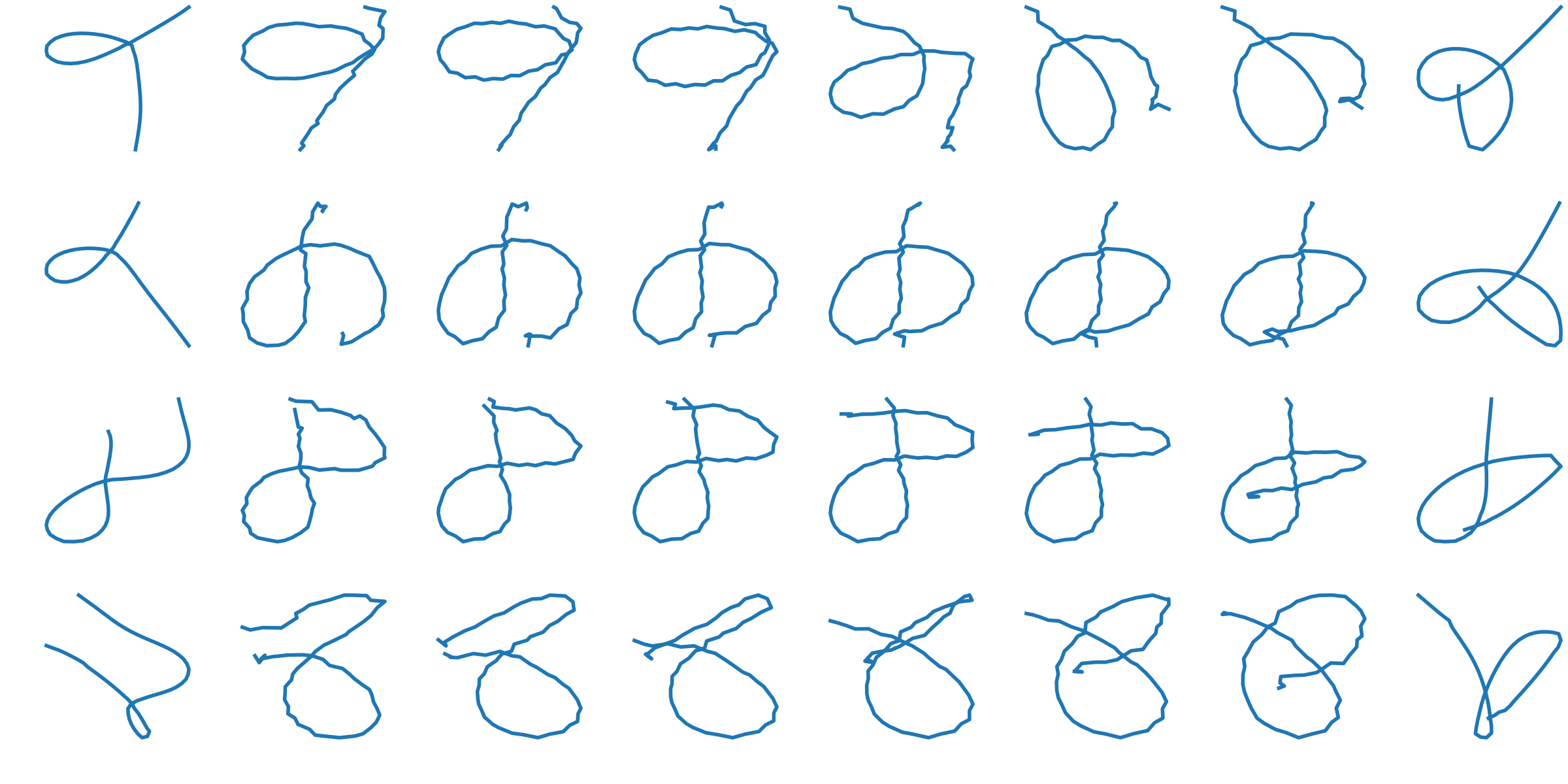}
    \caption{Visual plan generated by Causal InfoGAN. This model is trained on demonstration data for only one topological action, C(0,1,1). The plans are visually plausible, although the bottom two lines' motion strategy are different from the demonstration data.}
    \label{fig:GAN-single-stage}
\end{figure}

We show visualization of the generated plans from the single-stage model in Fig.~\ref{fig:GAN-single-stage}. The top two rows are from the training data and the bottom two rows are from the same evaluation set of rope configurations used in previous experiments. Within each row, the first and last images are the given start and end configurations, and the second and second last are the projected configurations. Different from our method, Causal InfoGAN requires the end configuration to be given, and we guarantee there is a feasible path for the start and end configurations shown in Fig.~\ref{fig:GAN-single-stage}. The distances from the given start and end configurations to their projected configurations are quite large. In the first row of Fig.~\ref{fig:GAN-single-stage}, the projected end configuration has a different topological state than the given one. Between the projected start and end configurations, the generated visual plans look plausible, although the bottom two rows adopted a motion strategy different from demonstration. The manipulation progress also seems to be uneven, e.g. middle columns in the first row have larger shape changes.

%\begin{figure}
%    \centering
%    \includegraphics[width=0.45\textwidth]{}
%    \caption{Visual plan generated by the multi-stage Causal InfoGAN on the training set. }
%    \label{fig:GAN-multi-stage}
%\end{figure}

\subsection{Plan execution on real robot} We show in the supplementary video that plans generated by our motion primitives can be executed on the real robot, with the help of visual tracking and MPC. Snapshots during the execution process are shown in the bottom row of Fig.~\ref{fig:teaser}.

\subsection{Generalization to more complex tasks}
\begin{figure}
    \centering
    \includegraphics[width=0.4\textwidth]{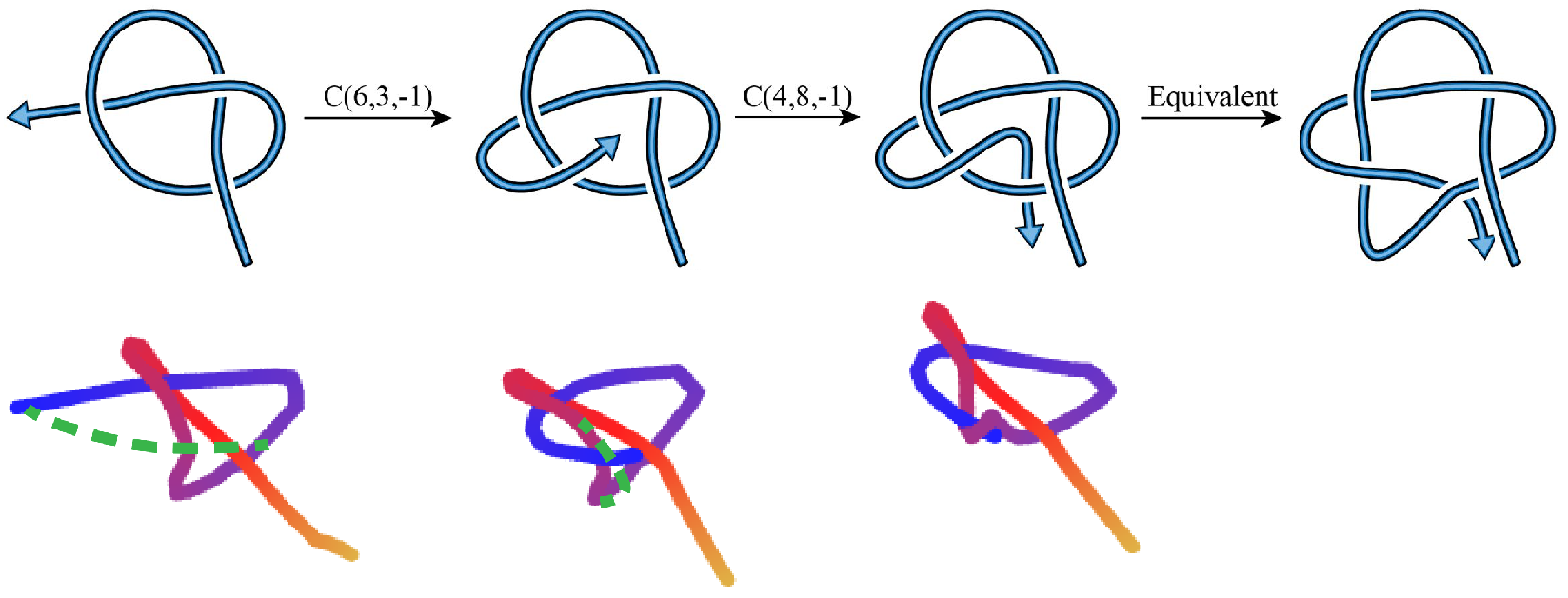}
    \caption{Topological path to a pentagram-like knot (top), and corresponding motion plan (bottom). Rope configurations are shown with color gradients, with blue as heads and orange as tails. Robot motion splines are shown in green.}
    \label{fig:demo-complex}
\end{figure}

We show that the topological motion primitives trained with the 8 topological actions above, where no more than three intersections are present in the rope states, can be used to find a plan for the more complex task of tying a pentagram-like knot starting from the overhand knot. The rope configurations (rainbow-colored solid lines) and the predicted robot motion splines (green dashed lines) are visualized in Fig.~\ref{fig:demo-complex}. This plan is found when only 12 branches are evaluated in the tree. We remark that the start configuration for this more complex task is different from the ones generated in the overhand knotting task, e.g. the one shown on the bottom-right corner of Fig.~\ref{fig:teaser}. We manually pulled out the head segment in order to demonstrate the more complex pentagram-like knot task. The trained topological motion primitives have not learned such pulling behavior, since such behaviors are not required in the overhand knot and not present in the demonstration data. We leave this to future work.

\vspace{-5pt}
\section{Conclusion}
We propose a 3-level motion planning and control algorithm for knot tying. At the top level, knot theory decomposes knotting tasks into sequences of abstract topological actions. The set of topological actions is shared across all knotting tasks. Our key contribution is at the second level, which is a set of topological motion primitives that generates robot motion trajectories for each topological action, conditioned on the observed rope configurations. At the bottom level, the predicted robot motion trajectories are simulated, and MPC tracks the resulting rope trajectory on real robots. We train the topological motion primitives by imitation and reinforcement learning. To generalize human demonstrations of simple knots into more complex knots, we observe similarities in the motion strategies of different topological actions, and designed the network architecture accordingly to learn such similarities. We demonstrate that our learned motion primitives have significantly higher success rate in completing the required topological actions, compared to baseline methods. When the motion primitives are used as samplers in a tree-search framework, a much smaller tree is grown before a valid motion plan is found. We verified that the generated robot motion plan can be executed on a real robot using visual tracking and MPC. We also demonstrated that our learned motion primitives can plan for a more complex pentagram-like knot, even though the human demonstrations are only on simpler tasks. This suggest that it is possible to scale the learned motion primitives to increasingly more complex knotting tasks without human intervention.

There may be several challenges in extending this work to a wide range of object types and knot types. Although we are confident this method applies to ropes with varying thickness and length, other object types such as cables, may be very stiff such that they return to their original shapes when released from the gripper. Very thin threads such as surgical suture may be hard to observe from depth images thus making 3D perception hard. To extend to more complex knot types, the robot would need to learn additional skills, such as pulling one segment from underneath existing intersections, or rearranging the rope to make next steps easier. The former may also require two robot arms to cooperate. These challenges may be addressed by some changes to the method. For example, by allowing each motion primitive to predict more than one robot motion spline when necessary, and providing demonstrations of re-arranging or pulling. To use multiple arms, the actions space may be extended. For example to acquire the pulling skill, the motion primitive could additionally predict an end-effector pose for the second arm to hold one point of the rope, while the first arm pulls another segment.

\footnotesize{
\bibliographystyle{IEEEtranN}
\bibliography{IEEEabrv,ref}
}
\end{document}